# Consensus and Consistency Level Optimization of Fuzzy Preference Relation: A Soft Computing Approach

Sujit Das
Dept. of Comp. Sc. & Engg., Dr. B.C. Roy Engg. College,
Durgapur, India
sujit_cse@yahoo.com

Samarjit Kar
Dept. of Mathematics, National Institute of Technology
Durgapur, India
kar_s_k@yahoo.com

*Abstract* - **In group decision making (GDM) problems fuzzy preference relations (FPR) are widely used for representing decision makers' opinions on the set of alternatives. In order to avoid misleading solutions, the study of consistency and consensus has become a very important aspect. This article presents a simulated annealing (SA) based soft computing approach to optimize the consistency/consensus level (CCL) of a complete fuzzy preference relation in order to solve a GDM problem. Consistency level indicates an expert's preference quality and consensus level measures the degree of agreement among experts' opinions. This study also suggests the set of experts for the necessary modifications in their prescribed preference structures without intervention of any moderator.**

*Keywords - Consistency/Consensus level, fuzzy preference relation, group decision making, simulated annealing.*

## I. INTRODUCTION

Consistency level along with consensus degree have a vital role in the decision making process. Optimization technique deals with selecting the best possible alternatives among a set of feasible alternatives. Optimization of CCL level is mandatory in order to achieve a quality decision. GDM process consists of a team of decision makers who provide their preferences over a set of alternatives. Multiplicative preference relation and fuzzy preference relation are two main types of preferences relations. Saaty [1] in 1980 and Wang and Xu [2] in 1990 introduced multiplicative preference relation which was defined as a reciprocal matrix. In 1978, Orlovsky [3] proposed FPR using complementary matrix to show an expert's opinion in decision making process. Chiclana et al. [4, 5] used reciprocal relations as uniform preference to present preference orderings and explored fuzzy multipurpose decision making problems. Herrera-Viedma et al. [6-8] proposed some GDM models using incomplete FPRs. Das et al. [12] have proposed a decision making procedure using FPR. Recently, a number of research works [13-22] have been published on decision making using soft computing techniques.

The aim of this article is to optimize the CCL level of FPR using SA [9] to achieve improved GDM process. Consensus degree measurement is used to evaluate the degree of agreement of all experts and a consistency level measurement is used to identify the quality intensity of each expert's opinion. To calculate the CCL, initially all missing values of all the incomplete FPRs are estimated using consistency based estimation procedure [8]. Then the consistency measure of each expert is computed and a global consistency level is measured. The method proposed by Herrera-Videma [7] has been used to compute the consensus degree of all the experts. The application of SA allows achieving consistent solutions with a high consensus degree without the intervention of any moderator. Fig. 1 shows the use of SA for GDM.

This article is organized as follows: Section II describes the various used methodologies. Simulated annealing algorithm is presented in section III. Numerical illustration is done in section IV followed by conclusion in section V.

## II. METHODOLOGY

### A. Fuzzy Preference Relation [8]

Among the different existing representation formats, majority of experts FPRs to express their opinions. This is one of the mostly used tools for GDM because of its effectiveness for modeling decision processes and easiness for aggregating experts' preferences into group ones [6].

*Definition 1.* A fuzzy preference relation $P = (p_{ij})_{n \times n}$ on a set of alternatives $X$ is a fuzzy set on the product set $X \times X$, i.e. it is characterized by a membership function $\mu_p : X \times X \to [0,1]$. Every value in the matrix $P$ represents the preference degree or intensity of preference of the alternative $x_i$ over $x_j$. $p_{ij} = 1/2$ indicates indifference between $x_i$ and $x_j$ $(x_i \sim x_j)$. $p_{ij} = 1$ indicates that $x_i$ is absolutely preferred to $x_j$. $p_{ij} > 1/2$ indicates that $x_i$ is preferred to $x_j$ $(x_i \succ x_j)$. Based on this interpretation, it can be said $p_{ii} = 1/2 \ \forall i \in \{1,....,n\} (x_i \sim x_i)$.

*Definition 2.* A function $f : X \to Y$ is partial when not every element in the set X necessarily maps onto an element in the set Y. When every element from the set X maps onto one element of the set Y then we have a total function.

*Definition 3.* An incomplete fuzzy preference relation P on a set of alternatives X is a fuzzy set on the product set X×X that is characterized by a partial membership function.

*Definition 4.* Additive transitivity for fuzzy preference relations can be seen as the parallel concept of Saaty's consistency property for multiplicative preference relations [10]. The mathematical formulation of the additive transitivity was given by Tanino in [11].

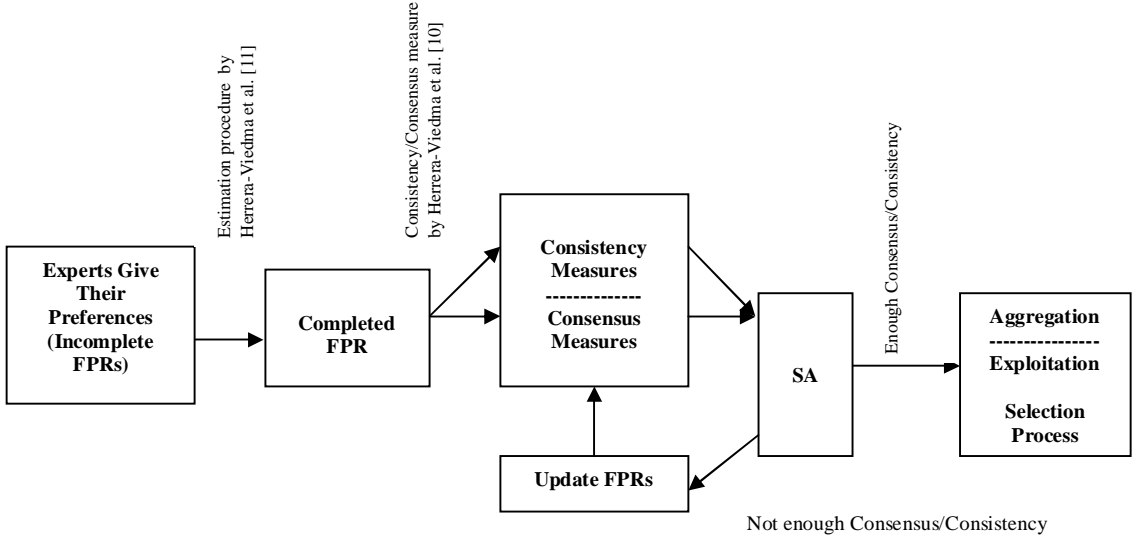

Figure 1. Use of SA for GDM with incomplete FPR

$$(p_{ij} - 0.5) + (p_{jk} - 0.5) = (p_{ik} - 0.5), \forall i, j, k \in \{1,...,n\} \quad (1)$$

Expression (1) can be rewritten as

$$p_{ik} = p_{ij} + p_{jk} - 0.5, \forall i, j, k \in \{1,....,n\}. \quad (2)$$

Additive transitivity implies additive reciprocity, because $p_{ii} = 0.5 \; \forall i$ if we make $k = i$ in (1), then we have $p_{ij} + p_{ji} = 1 \forall i, j \in \{1,...,n\}$.

### B. Estimation of Missing Value [7]

The preference value $p_{ik} (i \neq k)$ can be estimated using an intermediate alternative $x_j$ in three different ways. From $p_{\{ik\}} = p_{\{ij\}} + p_{\{jk\}} - 0.5$, it is found that the estimate $cp_{\{ik\}}^{\{j1\}} = p_{\{ij\}} + p_{\{jk\}} - 0.5$. From $p_{\{jk\}} = p_{\{ji\}} + p_{\{ik\}} - 0.5$, it is found that the estimate $cp_{\{ik\}}^{\{j2\}} = p_{\{jk\}} - p_{\{ji\}} + 0.5$. From $p_{\{ij\}} = p_{\{ik\}} + p_{\{kj\}} - 0.5$, It is found that the estimate $cp_{\{ik\}}^{\{j3\}} = p_{\{ij\}} - p_{\{kj\}} + 0.5$. The overall estimated value $cp_{ik}$ of $p_{ik}$ is obtained as the average of all possible $cp_{\{ik\}}^{\{j1\}}$, $cp_{\{ik\}}^{\{j2\}}$ and $cp_{\{ik\}}^{\{j3\}}$ values as given below.

$$cp_{ik} = \frac{\sum_{j=1; i \neq k \neq j}^{n} cp_{\{ik\}}^{\{j1\}} + cp_{\{ik\}}^{\{j2\}} + cp_{\{ik\}}^{\{j3\}}}{3(n-2)}.$$

### C. Finding Consistency Level [7,8]

The consistency level associated with a preference value $p_{\{ik\}}$ is defined as $CL_{\{ik\}} = (1 - \varepsilon p_{\{ik\}})$.

When investigation is done with complete fuzzy preference relation, for finding its consistency level, one has to calculate the value of average deviations of all $n-2$ possible estimates $cp_{\{ik\}}^{\{jl\}} (l \in \{1,2,3\})$ with respect to the actual value $p_{\{ik\}}$. The following expressions are used to find out the deviations:

$$\varepsilon p_{\{ik\}}^{\{1\}} = \frac{\sum_{j=1, j \neq i, k}^{n} \left| cp_{\{ik\}}^{\{j1\}} - p_{\{ik\}} \right|}{n-2}, \quad \varepsilon p_{\{ik\}}^{\{2\}} = \frac{\sum_{j=1, j \neq i, k}^{n} \left| cp_{\{ik\}}^{\{j2\}} - p_{\{ik\}} \right|}{n-2}$$

$$\varepsilon p_{\{ik\}}^{\{3\}} = \frac{\sum_{j=1, j \neq i, k}^{n} \left| cp_{\{ik\}}^{\{j3\}} - p_{\{ik\}} \right|}{n-2}$$

When the information provided in a fuzzy preference relation is completely consistent, then all $cp_{\{ik\}}^{\{jl\}} \in [0,1] (l \in \{1,2,3\}; \forall j \in \{1,...,n\}$ coincide with $p_{\{ik\}}$. The following expression (11) is used to measure the error in [0, 1] expressed in a preference degree between two alternatives.

$$\varepsilon p_{\{ik\}} = \frac{2}{3} \cdot \frac{\varepsilon p_{\{ik\}}^{\{1\}} + \varepsilon p_{\{ik\}}^{\{2\}} + \varepsilon p_{\{ik\}}^{\{3\}}}{3}. \quad (11)$$

So $\varepsilon p_{\{ik\}}$ can be used to find out the consistency level $CL_{\{ik\}}$ between the preference degree $p_{\{ik\}}$ and the rest of the preference values of the fuzzy preference relation.

When $CL_{\{ik\}} = 1$, then $\varepsilon p_{\{ik\}} = 0$, so there is no inconsistency at all. The lower the value of $CL_{\{ik\}}$, the higher the value of $\varepsilon p_{\{ik\}}$ and the more inconsistent is $p_{\{ik\}}$ with respect to the rest of information. The

consistency level of a fuzzy preference relation $p$ is defined as follows:

$$CL_p = \sum_{i,k=1, i \neq k}^{n} CL_{\{ik\}} \Big/ (n^2 - n).$$

with $CL_p \in [0,1]$. When $CL_p = 1$, the preference relation $P$ is fully consistent; otherwise, the lower $CL_p$ the more inconsistent $P$.

### D. Computing Consensus Measures [7]

Consensus degrees are used to measure the actual level of consensus in the process which can be identified at different levels of fuzzy preference relation like as pairs of alternatives, alternatives and relations. By measuring consensus degree one will be able to identify which experts are close to the consensus solution or in which alternatives the experts are having more trouble to reach consensus.

*1) Similarity matrix (SM):* For each pair of experts $(e_h, e_l)(h < l)$ a similarity matrix is defined as

$SM^{hl} = (sm_{\{ik\}}^{hl})$ where $sm_{\{ik\}}^{hl} = 1 - \left| p_{\{ik\}}^{-h} - p_{\{ik\}}^{-l} \right|$

Then a collective similarity matrix $SM = (sm_{\{ik\}})$ is obtained by aggregating all the $(m-1) \times (m-2)$ similarity matrices using arithmetic mean as the aggregation function.

*2) Consensus degree on pairs of alternatives:* The consensus degree on a pair of alternatives $(x_i, x_k)$, denoted by $cop_{\{ik\}}$, is defined to measure the consensus degree amongst all the experts on that pair of alternatives

$cop_{\{ik\}} = sm_{\{ik\}}$

*3) Consensus degree on alternatives:* The consensus degree on alternative $x_{\{i\}}$, denoted by $ca_{\{i\}}$ is defined to measure the consensus degree among all the experts on that alternative, defined below.

$$ca_{\{i\}} = \sum_{k=1, k \neq i}^{n} (cop_{\{ik\}} + cop_{\{ki\}}) \Big/ 2(n-1)$$

*4) Consensus degree on the relation:* The consensus degree on the relation, denoted by $CR$ is defined to measure the global consensus degree among all the experts' opinions

$$CR = \sum_{i=1}^{n} ca_{\{i\}} \Big/ n$$

### III. SIMULATED ANNEALING

#### A. Introduction

SA was introduced by Kirkpatrick [9] in 1983, inspired by annealing process in physics. The idea behind simulated annealing is based on the fact that a material is being heated and then slowly cooled down in a controlled manner, the mobility due to temperature of its particles is lost, and these particles reach a crystal structure along the solidification. The crystal state is the minimum energy free state of the system. The key of the process is the slow cooling, which allows a wide range of time for redistribution of the particles as they lose mobility. SA has proven to be a good technique to find solutions near the optimum (if not the optimum) within a reasonable computation time.

Simulated annealing is a stochastic search method based on the use of a local search. Local search iteratively replaces the current solution s by a new one s* until some stopping condition has been satisfied. The new solution which is termed as neighbor solution is created usually by a single move. The quality of the solution is characterized by its cost determined by a cost function. The goal of the search process is to minimize the cost function. But for SA, at any step, it either moves to a better neighbor solution if it finds one or to a worse solution with a certain probability. It includes the worse solution with an expectation that in near future it might be able to produce some better solutions.

#### B. Procedure

```
Procedure SA   /* detailed procedure*/
{input a trial solution S;    c = cost(S);    c* = c;
freezecount = 0; S*=S;
initialize temp, fastfactor, tempfactor;
initialize frzlim, sizefactor, minpercent, tcent;
while ( freezecount < frzlim )
{changes = trials = 0;
  while ( trials < sizefactor * N )
  {/* N is determined by the size of the problem*/
   trials = trials + 1;
   generate a random neighbour S' of S;
   c' = cost(S');  Δ = c'- c;
   if (S' is feasible and cost(S') < c* )
     { S* = S'; c* = cost(S');
     /* save best feasible solution found so far */ } if (Δ ≤ 0)
     { changes = changes + 1; c = c';S = S'; }
 /* downhill move */
  else
  {/* possible uphill move */
      choose a random number r in [0,1];
      if ( r <= exp(-Δ/temp) )
          { changes = changes+1; c = c'; S = S';}
   }
  }
if (changes/trials ≥ tcent )
   temp = fastfactor * temp;
   /* reduce temperature quickly */
else
   temp = tempfactor * temp;
   /* reduce temperature slowly */
 if ( changes/trials < minpercent )
   freezecount = freezecount+1;
else freezecount = 0;
}//end of procedure
```

#### C. NUMERICAL ILLUSTRATIONS

This article has used opinions of four experts $E = \{e_1, e_2, e_3, e_4\}$ over a set of four alternatives $X = \{x_1, x_2, x_3, x_4\}$ and these opinions are expressed using FPRs $(P_1, P_2, P_3, P_4)$ which are initially supposed to be incomplete as given below.

$$P_1 = \begin{pmatrix} - & 0.33 & 0.7 & 0.6 \\ x & - & x & x \\ x & x & - & x \\ x & x & x & - \end{pmatrix}, P_2 = \begin{pmatrix} - & x & 0.7 & 0.3 \\ 0.6 & - & x & 0.7 \\ 0.3 & x & - & x \\ x & 0.47 & x & - \end{pmatrix},$$

$$P_3 = \begin{pmatrix} - & 0.3 & 0.5 & 0.75 \\ 0.6 & - & x & 0.6 \\ x & 0.7 & - & x \\ 0.3 & 0.4 & x & - \end{pmatrix}, P_4 = \begin{pmatrix} - & x & 0.6 & 0.3 \\ 0.4 & - & 0.45 & 0.2 \\ 0.5 & 0.6 & - & 0.3 \\ 0.7 & 0.7 & 0.7 & - \end{pmatrix}.$$

Each incomplete FPR is completed by means of estimation procedure described by Herrera-Viedma et al. [8]. The corresponding completed fuzzy preference relations $CP_h, h = (1, 2, 3, 4)$ for $P_1, P_2, P_3$ and $P_4$ are respectively

$$CP_1 = \begin{pmatrix} - & 0.33 & 0.7 & 0.6 \\ 0.67 & - & 0.87 & 0.77 \\ 0.3 & 0.13 & - & 0.4 \\ 0.4 & 0.23 & 0.6 & - \end{pmatrix} CP_2 = \begin{pmatrix} - & 0.19 & 0.7 & 0.3 \\ 0.6 & - & 0.8 & 0.7 \\ 0.3 & 0.2 & - & 0.1 \\ 0.49 & 0.47 & 0.9 & - \end{pmatrix}$$

$$CP_3 = \begin{pmatrix} - & 0.3 & 0.5 & 0.75 \\ 0.6 & - & 0.65 & 0.6 \\ 0.85 & 0.7 & - & 0.78 \\ 0.3 & 0.4 & 0.25 & - \end{pmatrix} CP_4 = \begin{pmatrix} - & 0.59 & 0.6 & 0.3 \\ 0.4 & - & 0.45 & 0.2 \\ 0.5 & 0.6 & - & 0.3 \\ 0.7 & 0.7 & 0.7 & - \end{pmatrix}$$

Consistency measures of each expert $E = \{e_1, e_2, e_3, e_4\}$ are

$CL_1 = 1.0, CL_2 = 0.9, CL_3 = 0.87, CL_4 = 0.97$.

The global consistency level is

$CL = (1.0 + 0.9 + 0.87 + 0.97)/4 = 0.93$.

To find out the consensus degree six possible SMs between every pair of experts (not included for simplicity) are investigated and the collective one is

$$SM = \begin{pmatrix} - & 0.79 & 0.88 & 0.72 \\ 0.87 & - & 0.77 & 0.7 \\ 0.69 & 0.65 & - & 0.64 \\ 0.79 & 0.75 & 0.66 & - \end{pmatrix}$$

From $SM$ the consensus degree on relation (CR) is obtained as 0.74. The system has calculated CCL using $\delta = 0.65$ and the value obtained is

$CCL = (1 - 0.65) \times 0.93 + 0.65 \times 0.74 = 0.81$.

The SA procedure follows the mechanism mentioned in section III (B). The CCL is checked against a minimum threshold value, if $\gamma \geq 0.89$ the SA procedure ends. Otherwise it will check until the temperature level is reached below a minimum fixed level. When the CCL will be beyond the threshold limit, the suggested preference values $SPV_h, h = \{1, 2, 3, 4\}$ of all the experts are as follows.

$$SPV_1 = \begin{pmatrix} - & 0.03 & 0.50 & 0.50 \\ 0.97 & - & 0.77 & 0.07 \\ 0.50 & 0.23 & - & 0.10 \\ 0.50 & 0.93 & 0.90 & - \end{pmatrix}, SPV_2 = \begin{pmatrix} - & 0.09 & 0.90 & 0.40 \\ 0.70 & - & 0.80 & 0.20 \\ 0.10 & 0.20 & - & 0.10 \\ 0.39 & 0.97 & 0.90 & - \end{pmatrix},$$

$$SPV_3 = \begin{pmatrix} - & 0.00 & 0.90 & 0.65 \\ 0.90 & - & 0.85 & 0.10 \\ 0.45 & 0.50 & - & 0.18 \\ 0.40 & 0.90 & 0.85 & - \end{pmatrix} SPV_4 = \begin{pmatrix} - & 0.09 & 0.70 & 0.40 \\ 0.90 & - & 0.85 & 0.00 \\ 0.40 & 0.20 & - & 0.10 \\ 0.60 & 0.90 & 0.90 & - \end{pmatrix}$$

Here CL = 0.87, CR = 0.91 and CCL = 0.89. As experiment shows consistency level has decreased a little bit and consensus level has increased significantly. This occurs because the proposed system applies more importance on consensus criteria than the consistency one.

### D. Conclusion

This article has presented an optimization approach for GDM problems with incomplete FPRs guided by the additive consistency property. It has used SA for optimizing CCL of a set of FPRs provided by a set of experts. This method executes automatically without the participation of human operator and suggests the set of experts for required changes in their prescribed preference relations. This article is divided into mainly two phases: estimation of missing preference values and optimizing the CCL level. This procedure estimates the missing information in an expert's incomplete FPR using only the preference values provided by that particular expert. Future scope of this research work would be incrementing the CCL level without decrementing the consistency level.